\documentclass{book}    
\usepackage{piers}  
\usepackage[]{graphicx}
\graphicspath{{Figures/}}
\usepackage{subfigure}
\usepackage{amsmath}
\usepackage{amssymb}
\usepackage{epsfig}
\usepackage{cite}
\usepackage{color}
\usepackage{balance}
\usepackage{pgfplots}
\usepackage{wrapfig}
\usepackage{lipsum}
\usepackage{hyperref}

\begin{document}

\title{Analyzing Cloud Optical Properties Using Sky Cameras}
\maketitle

\begin{authors}

{\bf Shilpa Manandhar}$^{1}$ $^{*}$, {\bf Soumyabrata Dev}$^{2}$, {\bf Yee Hui Lee}$^{3}$, {\bf and Yu Song Meng}$^{4}$ $^{**}$\\
\medskip

$^{1}$Nanyang Technological University Singapore, Singapore 639798, email: \texttt{shilpa005@e.ntu.edu.sg}\\

$^{2}$Nanyang Technological University Singapore, Singapore 639798, email: \texttt{soumyabr001@e.ntu.edu.sg}\\

$^{3}$ Nanyang Technological University Singapore, Singapore 639798, email: \texttt{EYHLee@ntu.edu.sg}\\

$^{4}$National Metrology Centre, Agency for Science, Technology and Research (A$^{*}$STAR), Singapore 118221, email: \texttt{meng\_yusong@nmc.a-star.edu.sg}

$^{*}$ Presenting author\\
$^{**}$ Corresponding author

\end{authors}

\begin{paper}

\begin{piersabstract}
Clouds play a significant role in the fluctuation of solar radiation received by the earth's surface. It is important to study the various cloud properties, as it impacts the total solar irradiance falling on the earth's surface. One of such important optical properties of the cloud is the Cloud Optical Thickness (COT). It is defined with the amount of light that can pass through the clouds. The COT values are generally obtained from satellite images. However, satellite images have a low temporal- and spatial- resolutions; and are not suitable for study in applications as solar energy generation and forecasting. Therefore, ground-based sky cameras are now getting popular in such fields. In this paper, we analyze the cloud optical thickness value, from the ground-based sky cameras, and provide future research directions.

\end{piersabstract}

\psection{Introduction}

Cloud optical properties and cloud coverage contribute in rapid fluctuation of solar radiation during a day. A correct estimation of cloud properties is useful in the emerging field of solar energy harvesting and forecasting. Satellite images are one of the primary sources to study different cloud properties. The Moderate Resolution Imaging Spectroradiometers (MODIS) installed on National Aeronautics and Space Administration (NASA) Earth Observing System’s (EOS) Terra and Aqua satellites transmit visible and near-infrared spectral signatures~\cite{Wei_2004} to get the
information on cloud dynamics and properties like optical thickness, effective particle radius, particle phase, and liquid water content. It also provide cloud mask data, that can be correlated from ground observations~\cite{PIERS17a}. These instruments provide such important cloud properties information. However, they have poor temporal and spatial resolutions. Therefore, ground based sky cameras are now widely used to study the cloud properties~\cite{Dev2016GRSM}. In this paper, we obtain COT from MODIS satellite images, and analyze its relationship with the optical properties of visible-light images obtained from sky cameras. 
 
\psection{Data Collection}

We have installed a ground-based sky camera on a particular rooftop ($1.3483^\circ$ N, $103.6831^\circ$ E) of our university building at Nanyang Technological University Singapore. This sky camera is called WAHRSIS (Wide Angled High Resolution Sky Imaging System)~\cite{WAHRSIS,IGARSS2015a}. It captures the images of the sky scene at an interval of $2$ minutes. These images are captured in the visible-light range, and are useful to provide higher temporal and spatial information of the cloud information. Several events in the earth's atmosphere can be effectively studied using such ground-based cameras. These captured images also provide us an idea about the optical properties of the cloud, depending on the normalized luminance of the captured image. For this paper, we analyze the deviation of captured luminance from the clear-sky luminance~\cite{IGARSS_solar}, and correlate with COT.

The COT values are calculated from MODIS products, which has spatial resolution of 1 km$^{2}$ (1 pixel) and is available twice a day (4 UTC and 7 UTC). In this experiment, we process COT values for 3 km $\times$ 3 km area (9 pixels) with the location of sky camera as the center. The value of COT ranges from $0$ to $100$. A cloud optical thickness of value $100$ means that almost no light can pass through the clouds whereas a value of $0$ means the sky is relatively cloud free. 

\psection{Experiments \& Results}
For our experiments~\footnote{The source codes of these experiments are available online at \url{https://github.com/Soumyabrata/cloud-optical-thickness}.}, we consider all the MODIS observations in the year $2015$. We process the average COT values over the 9 pixels, from the extracted $3\times3$ cloud optical thickness matrix. This provides us an average COT value, corresponding to the MODIS timestamp.

Our sky cameras capture images at temporal resolutions of $2$ minutes, and therefore, has higher resolutions as compared to the MODIS satellite images. We compute the luminance of the circumsolar region, and normalize it with the camera parameters. This entire methodology of computing the luminance of the circumsolar region, from the sky cameras is described in our previous work~\cite{IGARSS_solar}. Based on this linear relationship between image luminance and solar irradiance, we can now compute the clear sky luminance. The clear sky luminance is referred as the corresponding luminance of the theoretical clear-sky model. We use the model by \cite{dazhi2012estimation}, to estimate the clear-sky irradiance, and subsequently the clear sky luminance. Finally, the luminance difference is the difference between the actual luminance and clear-sky luminance.

Taking the corresponding MODIS COT time as reference, we compute a $15$ minute average of luminance difference from sky images. We also normalize the values of the luminance difference and cloud optical thickness value. Figure~\ref{fig:COT-fig} shows the scatter plot of calculated average luminance difference and average COT values. 

\begin{figure}[htb]
\centering
\includegraphics[height=0.7\textwidth]{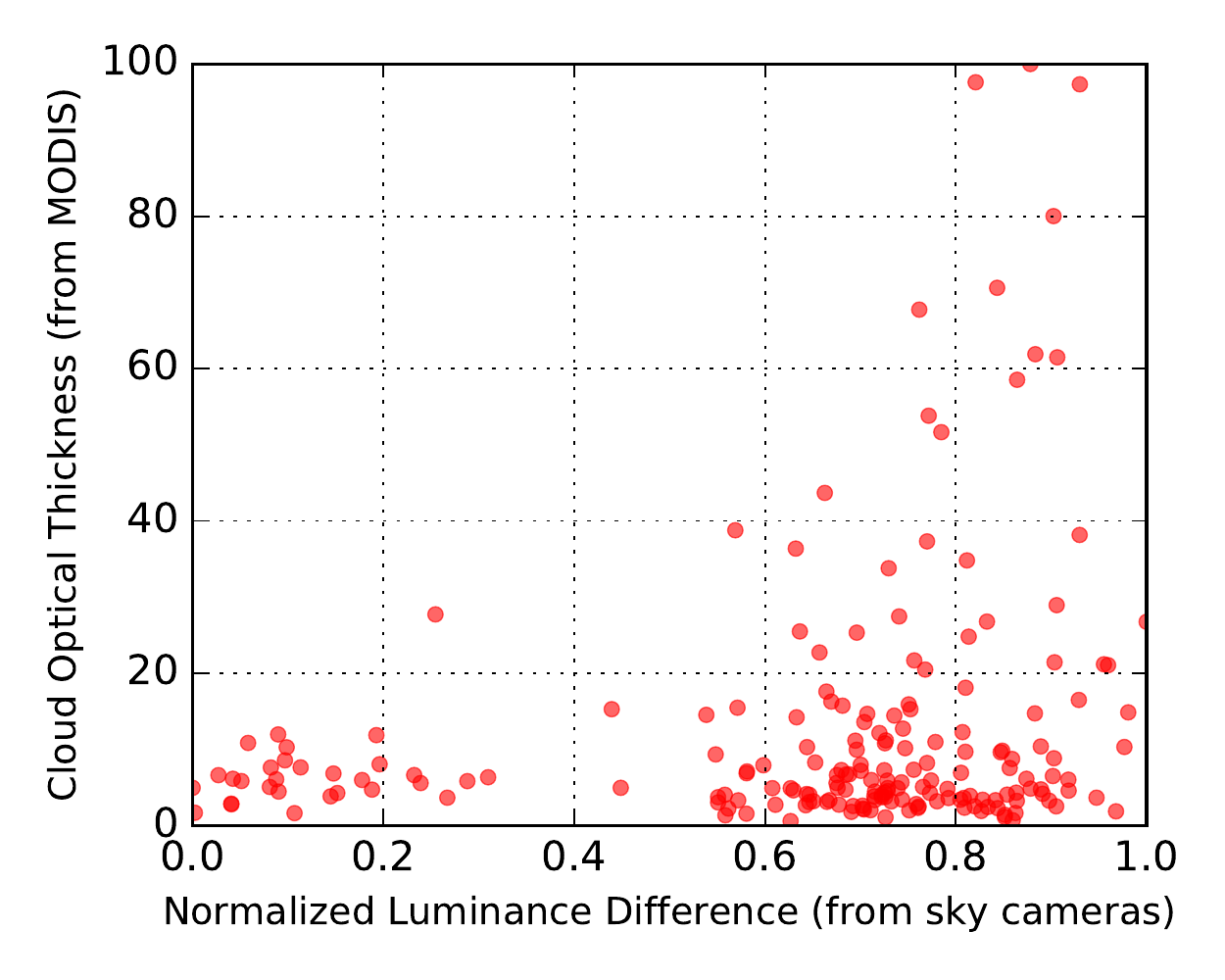}
\caption{Relation between cloud optical thickness (from MODIS) and normalized luminance difference (from ground-based sky camera).}
\label{fig:COT-fig}
\end{figure}

This is self explanatory -- high luminance difference reflects presence of clouds. And thus, cloud presence is reflected by a higher value of COT. Interestingly, we can also observe lower COT values at higher luminance difference values. This possible deviation can be explained because of the possible mismatch between the coverage of MODIS and sky cameras. Also, there can exist some time gap between MODIS time stamp and the sky imager time stamp. In the future, we plan to model the cloud optical thickness value directly from the normalized luminance difference of the sky cameras. This will help us to estimate COT values at much finer time resolutions.

\psection{Conclusion}
In this paper, we have studied the cloud optical thickness value using the ground-based sky cameras. This is an important contribution, as ground-based sky cameras, if installed at the experiment site, can provide higher time- and spatial- resolution of cloud formation. 
In our future work, we will propose a best fit line between COT and normalized luminance difference; and also work on designing Internet-of-Things (IoT) enabled wireless sensors~\cite{Deepu2017} for sky cameras. This will help the remote sensing analysts to compute COT values for finer time resolutions, as compared to the current standards.

\ack
The authors would like to thank Joseph Lemaitre for automatizing the acquisition and processing of MODIS multi bands images into a user-friendly framework.

\end{paper}

\end{document}